# RAPIDAI4EO: A CORPUS FOR HIGHER SPATIAL AND TEMPORAL REASONING

*Giovanni Marchisio[1], Patrick Helber[3], Benjamin Bischke[3], Timothy Davis[2], Caglar Senaras[2], Daniele Zanaga[4], Ruben Van De Kerchove[4], Annett Wania[2]*

[1]Planet Labs Inc., USA, [2]Planet Labs GmbH, Germany, [3]Vision Impulse GmbH and DFKI, Germany, [4]VITO NV, Belgium

**ABSTRACT**

Under the sponsorship of the European Union's Horizon 2020 program, RapidAI4EO will establish the foundations for the next generation of Copernicus Land Monitoring Service (CLMS) products. The project aims to provide intensified monitoring of Land Use (LU), Land Cover (LC), and LU change at a much higher level of detail and temporal cadence than it is possible today. Focus is on disentangling phenology from structural change and in providing critical training data to drive advancement in the Copernicus community and ecosystem well beyond the lifetime of this project. To this end we are creating the densest spatiotemporal training sets ever by fusing open satellite data with Planet imagery at as many as 500,000 patch locations over Europe and delivering high resolution daily time series at all locations. We plan to open source these datasets for the benefit of the entire remote sensing community.

## 1. INTRODUCTION

New catalogues of nearly daily or even intraday temporal data will soon dominate the global Earth Observation (EO) archives. However, there has been little exploration of artificial intelligence (AI) techniques to leverage the high cadence that is already possible to achieve through the fusion of multiscale, multimodal sensors. While the deluge of data makes it possible in theory to take a daily pulse of our planet, there has been little practical work on establishing foundational analytics and infrastructures that can leverage the temporal dimension at scale. A significant obstacle is that training data remains rare [1], [2], [3] relative to the spatiotemporal sampling which is necessary to adequately capture natural and man-made phenomenology latent in these large volumes of observations. And it is not easy to integrate all of the data that is available. Sensor interoperability issues and cross-calibration challenges present obstacles in realizing the full potential of these rich geospatial datasets.

Specific objectives of the project are to: 1) demonstrate the fusion of Copernicus high resolution satellite imagery and third party very high resolution imagery; 2) stimulate the development of analytics based on the availability of very high cadence imagery at a continental scale; 3) provide intensified monitoring of LU and LC with direct impact on the CORINE Land Cover (CLC) programme. CORINE is the flagship of CLMS and has been produced for the reference years of 1990, 2000, 2006, 2012 and 2018. New EU policy requirements call for an intensified monitoring of LC and of LU at a much higher level of detail and temporal cadence. The EIONET EAGLE group started addressing these issues by providing a framework to integrate geographical data from different sources and by disaggregating a landscape in its LC and LU components [4]. The next generation CLC, code-named CLC+, will provide a pan-European implementation of the EAGLE concept. The thematic content would be structured with separation of LC and LU for the needs of most application domains and complemented with characteristics for ecosystem mapping and assessment.

Our goal is to develop dedicated Deep Learning (DL) architectures to model the phenomenology inherent in high cadence observations. We want to drive a paradigm shift away from conventional approaches to change detection, which are typically based on measuring a few points in time, to a scenario that involves monitoring or tracking change continuously. Spatiotemporal explicit models trained on daily time series of multispectral images acquired over an entire year, or multiple years, could eventually learn how to disentangle phenology from structural change in a weakly or fully unsupervised mode.

## 2. DATA FUSION

We believe that the future of EO is in fusion, harmonization, and interoperability of satellite imagery. Planet has been pioneering a methodology, the CubeSat-Enabled Spatio-Temporal Enhancement Method (CESTEM) [5] [6] to enhance, harmonize, inter-calibrate, and fuse cross-sensor data streams leveraging rigorously calibrated 'gold standard' satellites (Sentinel, Landsat, MODIS) in synergy with superior resolution CubeSats from Planet. With reference to

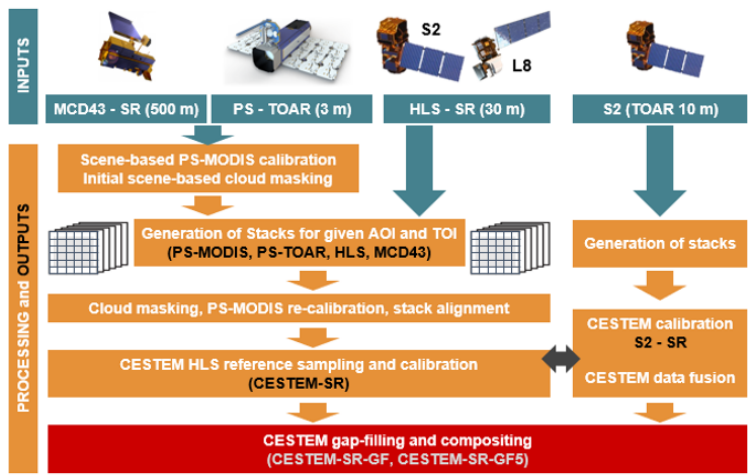

**Fig. 1:** Overview of CESTEM harmonization workflow

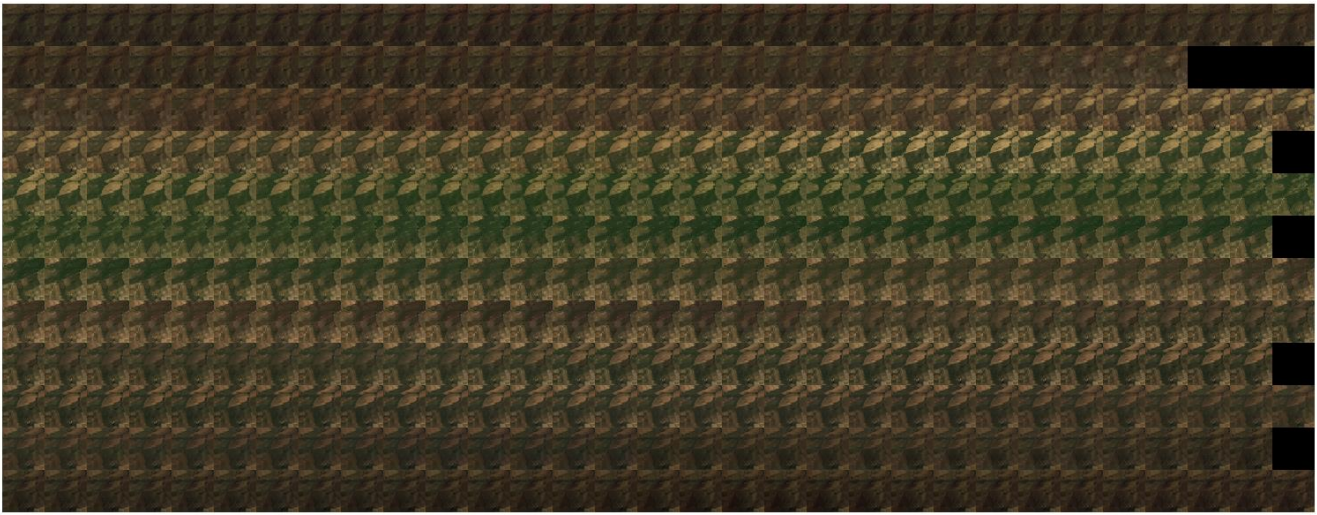

**Fig. 2:** Sample Fusion data cube from the RapidAI4EO corpus consisting of 1-year 3m, daily cloud-free imagery.

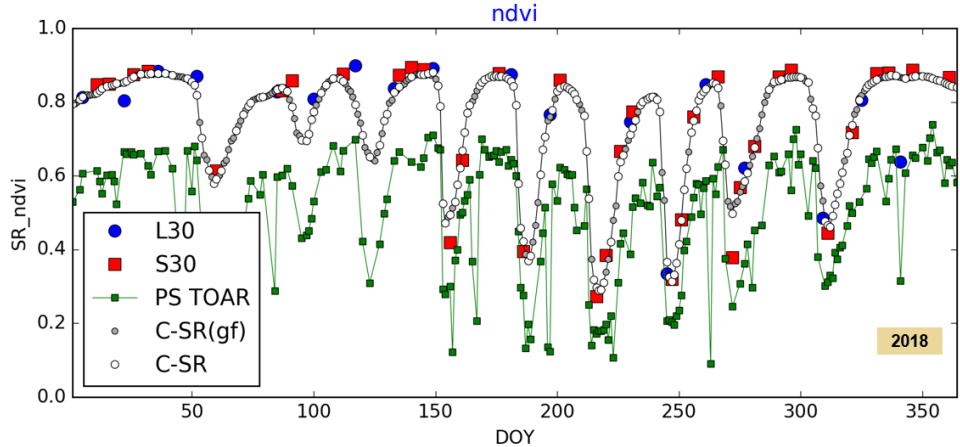

**Fig. 3:** The temporal sampling density and interoperability achievable with harmonized, daily gap-filled Fusion coverage (small white/gray circles) compared to Sentinel-2 (red squares) and Landsat measurements (blue circles).

Fig. 1, the steps involved are: 1) scene-based PS-MODIS calibration with initial scene-based cloud masking; 2) generation of stacks for given AOIs and TOIs, including PS-MODIS, PS-TOAR, HLS (Harmonized Landsat Sentinel), and MCD43; 3) cloud masking, PS-MODIS re-calibration, HLS-PS stack co-registration; 4) HLS reference sampling and calibration (for harmonized surface reflectance output); 5) multi-sensor fusion with gap filling and compositing (for daily or 5-day harmonized, high resolution surface reflectance output). The result is next generation L3H analysis ready data (ARD). This new, high cadence data stream called Fusion Monitoring delivers clean (i.e. free from clouds and shadows), gap-filled, daily, 3 m, temporally consistent, radiometrically robust, and sensor agnostic surface reflectance feeds synergizing inputs from both public and private sensor sources (Fig. 2 and 3)

## 3. ANNOTATION AND SAMPLING STRATEGY

The inspiration for the RapidAI4EO datasets comes from the recently released patch-based LULC EuroSAT [7] and BigEarthNet [8] corpora. The RapidAI4EO data sets bring together the best of both, addressing greater and improved sampling throughout all the countries in the EU and providing the temporal and seasonal variability necessary for improved modeling of intra-class variance.

Satellite imagery collected around the world from different geographic locations often contain large variations in the illumination and the appearance of LULC classes. This high intra-class variability can be a result of varying illumination conditions, climatic conditions, and phenology (e.g., at different times during the day and seasonal effects). Related research [9] shows the relative lack of geographic portability of classifiers even within the same country. In many cases, this is due to a strongly biased data distribution of the training dataset which leads to a poor generalization of the model. That is, the trained model overfits and the intra-class variability inherent in the underlying class is not covered. Therefore, in addition to spatial sampling, temporal sampling of LULC classes is of high importance for robust classification.

The RapidAI4EO corpus is based on the intersection of CORINE 2018 and coincident Sentinel-2 and Planet with all the countries in the EU, balancing relative country surface with relative LULC distribution to cover all climatic zones for each class. Patch annotations are multi-class and based initially on the CLC ontology, which currently includes 44 LC classes, and with an eye to supporting multiple ontologies and convergence with the Land Cover Classification System (LCCS). We sample 500,000 European locations with a patch size of 640m x 640m by partitioning locations into bins relative to each country surface distribution. Within these bins we randomly sample N locations and verify that the class distribution matches the original CORINE class distribution in each country, also ensuring that we have a minimum number of samples for each class. The higher spatial resolution of Planet offers the potential for modeling more LULC classes than possible at 10 m, but the greatest innovation is the addition of dense harmonized time series for both Sentinel-2 and Planet at all 500,000 patch locations for the entire year 2018. In the case of Planet Fusion, these consist of 365 (daily) time frames.

## 4. FUTURE MODEL EXPLOITATION

There is nothing that compares to dense temporal observations to create a robust historical baseline for every location on earth and enable robust change detection. The higher spatiotemporal resolution of the RapidAI4EO corpus should enable disambiguation of land covers across diverse climate zones, as well as an improved understanding of land use. The corpus provides the basis to test a variety of semi-supervised and fully unsupervised approaches to modeling. Underlaying all these is the principle of deep spatiotemporal encoding at the patch level.

Our previous work with the EuroSAT dataset and supervised patch classification using Convolutional Neural Networks shows 98% or better accuracy in classifying 10 parent LC classes derived from a 27 LULC ontology [7]. The hope in the new project is that we can go beyond this number of classes by exploiting the higher spatiotemporal resolution of our corpus.

At the opposite end of the spectrum are emerging DL architectures that can learn to separate information relating to seasonality from real, structural change in a semi or fully unsupervised way. They can potentially learn a robust historical baseline for every patch location on earth and enable robust change detection from dense observations. There are several competing approaches for extracting optimal feature representations for satellite images, some of which have been developed in the context of visual search and retrieval [10]. One simple unsupervised approach is to compare the newly extracted representation for daily collects with the observed statistical norm over time at any given patch location. This unsupervised approach to change detection will only improve with time as we get more confident with more images.

An interesting evolution on this approach is to split the latent feature representation in two parts: one part that represents information related to seasonality, illumination (part A), and one part that represents inherent structural information related to LULC (part B). The total loss function can be defined to minimize the reconstruction loss in adjacent images, and additionally minimize the difference in part B between them (assuming no structural changes have occurred at this location). Part A is left unconstrained to allow the network to represent differences due to seasonality, illumination, etc. by this part of the latent vector. This relies on the assumption that for the vast majority of randomly extracted baselines, there is no change between the different acquisition times except for seasonality or illumination. The idea of disentangling latent representations is taken further one step further in [11].

## 5. CONCLUSION

We believe that it is possible to close the growing gap between data availability and data interpretability in Earth Observations. The emergence of powerful new artificial intelligence methodologies must be matched by tools that enable the remote sensing community to exploit their full potential when applied to the anticipated increasing volumes, cadence, and diversity of sensor data. As it has been proved in other fields where artificial intelligence is revolutionizing entire industries (e.g., the self-driving car, video processing, voice recognition, face recognition), the availability of large volumes of training data representative of the phenomenology to be modeled is key to overcoming a fundamental obstacle.

In addition to providing a very large, high quality, high density, harmonized spatiotemporal training corpus, RapidAI4EO will inspire the next generation of Copernicus Land Monitoring applications by demonstrating large scale thematic change detection at a monthly cadence over all of Europe and show how this has a direct impact on improving the maintenance and quality of the CLC product. If successful, this framework would take us one step closer to measuring the global indicators for the Sustainable Development Goals. We believe that the future of EO is in fusion, harmonization, and interoperability of satellite imagery. Intensified monitoring leads to better understanding of land use and reduction of maintenance costs for all LC products.

## 6. ACKNOWLEDGEMENTS

This project has received funding from the European Union's Horizon 2020 research and innovation programme under grant agreement No 101004356.

## 7. REFERENCES

[1] M. Schmitt, J. Prexl, P. Ebel, L. Liebel, and X.X. Zhu, "Weakly Supervised Semantic Segmentation of Satellite Images for Land Cover Mapping -- Challenges and Opportunities", *arXiv preprint arXiv:2002.08254*, 2020

[2] G. Nayak, R. Ghosh, X. Jia, V. Mithal, and V. Kumar, "Semi-supervised Classification using Attention-based Regularization on Coarse-resolution Data", *arXiv preprint arXiv:2001.00994,* 2020

[3] N. Bengana and J. Heikkilä, "Improving land cover segmentation across satellites using domain adaptation", *arXiv preprint arXiv:1912.05000*, 2020

[4] S. Arnold, B. Kosztra, G. Banko, G. Smith, G.W. Hazeu, M. Bock, and S. Valcarcel Sanz,,(2013). "The EAGLE concept - A vision of a future European Land Monitoring Framework. ", *33rd EARSeL symposium Towards Horizon 2020*, 2013, Matera, Italy, http://sia.eionet.europa.eu/EAGLE/Outcomes/EARSeL-Symposium-2013_10_2_EAGLE-concept_Arnold-et-al.pdf

[5] R. Houborg and M. F.McCabe, "Impacts of dust aerosol and adjacency effects on the accuracy of Landsat 8 and RapidEye surface reflectances*", Remote Sensing of Environment*, vol. 194, 1 June 2017, Pages 127-145

[6] R. Houborg and M. F.McCabe, "A Cubesat enabled Spatio-Temporal Enhancement Method (CESTEM) utilizing Planet, Landsat and MODIS data", *Remote Sensing of Environment*, vol. 209, May 2018, Pages 211-226

[7] P. Helber, B. Bischke, A. Dengel and D. Borth, "EuroSAT: A Novel Dataset and Deep Learning Benchmark for Land Use and Land Cover Classification," *in IEEE Journal of Selected Topics in Applied Earth Observations and Remote Sensing*, vol. 12, no. 7, pp. 2217-2226, July 2019, doi: 10.1109/JSTARS.2019.2918242.

[8] G. Sumbul, M. Charfuelan, B. Demir and V. Markl, "Bigearthnet: A Large-Scale Benchmark Archive for Remote Sensing Image Understanding," *IGARSS 2019 - 2019 IEEE International Geoscience and Remote Sensing Symposium*, Yokohama, Japan, 2019, pp. 5901-5904, doi: 10.1109/IGARSS.2019.8900532.

[9] E. Maggiori, Y. Tarabalka, G. Charpiat and P. Alliez, "Can semantic labeling methods generalize to any city? the inria aerial image labeling benchmark," *2017 IEEE International Geoscience and Remote Sensing Symposium (IGARSS)*, Fort Worth, TX, 2017, pp. 3226-3229, doi: 10.1109/IGARSS.2017.8127684.

[10] R. Keislera,, S.W. Skillmana , S. Gonnabathulaa , J. Poehnelta , X. Rudelisa , and M.S. Warren, "Visual search over billions of aerial and satellite images", *Computer Vision and Image Understanding*, *arXiv preprint arXiv:2002.02624,* 2020

[11] E.H. Sanchez, M.Serrurier and M. Ortner," Learning Disentangled Representations of Satellite Image Time Series", *arXiv preprint arXiv:1903.08863,* 2019